\documentclass[twocolumn,10pt]{asme2e}

\usepackage{graphics}
\usepackage{graphicx}
\usepackage{amsmath}

\confshortname{IDETC/CIE 2017}
\conffullname{the ASME 2017 International Design Engineering Technical Conferences \&\\
              Computers and Information in Engineering Conference}

\confdate{August 6-9, 2017}
\confyear{2017}
\confcity{Cleveland}
\confcountry{USA}

\papernum{DETC2017/MR-67087}

\title{Kinematics and Workspace Analysis of a 3PPPS Parallel Robot with U-Shaped Base}

\author{Damien Chablat
    \affiliation{
	Laboratoire des Sciences du Num\'erique de Nantes\\
	UMR CNRS 6004\\
	44321 Nantes, France 95616\\
    Email: damien.chablat@cnrs.fr }	
}

\author{Luc Baron, Ranjan Jha     
    \affiliation{Mechanical Engineering Department, \\
 \'Ecole Polyt\'echnique de Montr\'eal, \\
 H3C 3A7 Qu\'ebec, Canada\\
	Email: [Luc.Baron, Ranjan.Jha]@polymtl.ca}
}

\begin{document}

\maketitle    

\begin{abstract}
{\it This paper presents the kinematic analysis of the 3-\underline{PP}PS parallel robot with an equilateral mobile platform and a U-shape base. The proposed design and appropriate selection of parameters allow to formulate simpler direct and inverse kinematics for the manipulator under study. The parallel singularities associated with the manipulator depend only on the orientation of the end-effector, and thus depend only on the orientation of the end –effector. The quaternion parameters are used to represent the aspects, i.e. the singularity free regions of the workspace. A cylindrical algebraic decomposition is used to characterize the workspace and joint space with a low number of cells. The discriminant variety is obtained to describe the boundaries of each cell. With these simplifications, the 3-\underline{PP}PS parallel robot with proposed design can be claimed as the simplest 6 DOF robot, which further makes it useful for the industrial applications.} \end{abstract}
\section{Introduction}
The architectures of robot manipulators can be classified based on the type of the kinematic chains connecting the output link of the manipulator to the base link, i.e., serial, parallel and hybrid architecture. In the serial architecture, the output link is connected to the base link by a single open loop kinematic chain. The kinematic chain is composed from a group of rigid links where each pair of adjacent links are interconnected by an active kinematic pair. Serial robots feature a large workspace volume and high dexterity, but suffer from several inherent disadvantages. The letters include low payload-to-weight ratio, poor force exertion capability and low precision. A parallel robot is a mechanical system with a closed-loop kinematic chain whose end-effector is linked to the base by several independent kinematic chains. Parallel robots can be categorized in two different types as fully parallel and non-fully parallel manipulators based on the relation between the number of chains and the degree of freedom of the end-effector. Parallel architectures provide high rigidity and high payload-to-weight ratio, high accuracy, low inertia of moving parts, high agility, and simple solution for the inverse kinematics problem (IKP). The fact, the load is shared by several kinematic chains in a high payload-to-weight ratio and rigidity. The disadvantages are the limited work volume, low dexterity, complicated direct kinematics solution, and singularities that occur both inside and on the envelope of the workspace volume.

Most examples of 6-DOF fully-parallel manipulators may be classified by the type of their six identical serial chains being UPS \cite{Merlet:2006,Pierrot:1998,Corbel:2008,sto:1993,ji:1999}, RUS \cite{hon:1997,mer:1991}, or PUS \cite{hun:1983}. Independently from the type of kinematic chains, there exists three legged robots \cite{ali:1994,behi:1988}, with only three legs and two actuators per leg and decoupled robots \cite{jin:2004,lal:1997}, in which the translational and rotational degrees of freedom of the mobile platform are decoupled. The first implementation of such parallel architecture by \cite{gough:1962} presented a six degrees of freedom tire test machine with base and moving platforms interconnected by six extensible screw jacks. Stewart presented a parallel robot for a six-degrees of freedom flight simulator \cite{stewart:1965}. This robot was composed of a base and a triangular moving platform with three extensible links connecting the moving platform to the base. Recently, a six-legged parallel robot was introduced in \cite{Seward:2014} with simpler direct kinematics problem (DKP) which can be solved easily by partitioning the orientation and the position of the mobile platform. However, the workspace size is limited for orientation due to the interferences between the legs.

A six DOF epicyclic parallel manipulator, Monash Epicyclic-Parallel Manipulator (MEPaM), is presented in \cite{Chen:2012} with all actuators mounted on the base, parallel singularity is independent on the position of the end-effector. Non-singular assembly mode changing of a six DOF parallel manipulator 3-\underline{PP}PS manipulator is shown in \cite{Caro:2012}. Eight solutions to its DKP, several assembly modes can be connected by a non singular trajectories by encircle the cusp points in the joint space. The singularity analysis of a six dof three-legged parallel manipulator for force feedback interface, using Grassmann-Cayley algebra jacobian and Gr\"{o}bner basis \cite{Faugere:1999}, shown in \cite{Caro:2010}. The cylindrical algebraic decomposition (CAD) algorithm \cite{Cbook:75} is used to study the workspace and joint space, and a Gr\"{o}bner based elimination process is used to compute the parallel singularities of the manipulator \cite{jha:2016,chab:2014,Jhaa:2016}.

This paper presents the kinematic analysis of the 3-\underline{PP}PS parallel robot derived from \cite{Chen:2012} with an equilateral mobile platform and a U-shape base. The parallel singularities associated with the manipulator are independent of the position of the end-effector, while depends only on the orientation of the end –effector. The quaternion parameters are used to represent the aspects, i.e. the singularity free regions of the workspace. A CAD is used to characterize the workspace and joint space with a low number of cells. The discriminant variety is obtained to describe the boundaries of each cell \cite{Lazart:2007}. 

The outline of this paper is as follows. Section 2 describes the architecture of the manipulator, including constraint equations associated with the manipulator. Section 3 discusses the computation of parallel singularities. Section 4 and 5 formulates the direct kinematics and inverse kinematics problems for the mechanism under study. Section 6 presents the workspace and joint space analysis of 3-\underline{PP}PS parallel robot. Section 7 finally concludes the paper.
\section{Mechanism Architecture}
The robot under study is based on the MEPaM, developed at the Monash University \cite{Chen:2012,Caro:2010}. This architecture is derived from the 3-\underline{PP}SP introduced earlier in \cite{Byun:1997}. In the original design, the first actuators of each leg  are in orthogonal directions. With this design, the robot admits up to six solutions to the DKP and it is able to perform non-singular assembly mode-changing trajectories  \cite{Caro:2012}. The main property of this robot is that the parallel singularity postures depends only on  orientation of the end-effector. Another design was introduced in \cite{Chen:2012} where the two first actuated joints are on the faces of a prism.  This design is simpler from the kinematic point of view. The singular configurations are ease to define in the orientation space but the workspace and joint space needs to be analyzed  in a 5 or 6 dimensional space.

The new design of the 3-\underline{PP}PS robot is derived from \cite{Bai:2009}, where the authors investigate a 3-PPS planar parallel robot with the actuated prismatic joints placed in a U-shaped base.
\subsection{Geometric Parameters}
\begin{figure*}[!ht]
\begin{center}
\includegraphics[scale=.35]{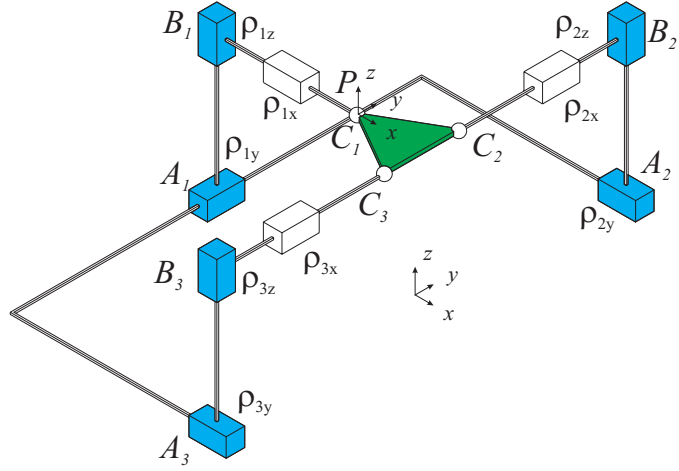}
\end{center}
\caption{The 3-\underline{PP}PS parallel robot and its parameters in its ``home'' pose with the actuated prismatic joints in blue, the passe joints in white and the mobile platform in green}
\label{Fig:Robot}
\end{figure*}

The three legs are identical and made with two actuated prismatic joints plus one passive prismatic joint and a spherical joint (Figure~\ref{Fig:Robot}). The axes of first three joints form an orthogonal reference frame.

The coordinates of the point $C_1$ are $\rho_{1x}$, $\rho_{1y}$ and $\rho_{1z}$, where the last two are actuated. For the sake of the study, we have defined an origin $A_i$ for each leg as
\begin{eqnarray}
 {\bf A}_1&=&[ 2, \rho_{1y}, \rho_{1z}]^T \\
 {\bf A}_2&=&[-\rho_{2y}, 2, \rho_{2z}]^T \\
 {\bf A}_3&=&[ \rho_{3y},-2, \rho_{3z}]^T
\end{eqnarray}
The coordinates of  $C_2$ and $C_3$ are obtained by a rotation around the $z$ axis by $\pi/2$ and $-\pi/2$, respectively. 
\begin{eqnarray}
 {\bf C}_1&=&[ \rho_{1x}, \rho_{1y}, \rho_{1z}]^T \\
 {\bf C}_2&=&[-\rho_{2y}, \rho_{2x}, \rho_{2z}]^T \\
 {\bf C}_3&=&[ \rho_{3y},-\rho_{3x}, \rho_{3z}]^T
\end{eqnarray}
There are several ways to attach the moving frame to the mobile platform. Below, we present three locations of the origin on the mobile equilateral platform.

{\bf Location 1: Center of the mobile platform}
\begin{eqnarray}
{\bf V}_1 &=& [ \sqrt{3}/3,0,0]^T \\
{\bf V}_2 &=& [-\sqrt{3}/6,1/2,0]^T \\
{\bf V}_3 &=& [-\sqrt{3}/6,-1/2,0]^T
\end{eqnarray}

{\bf Location 2: One corner for the origin and one other of a reference for the angles}
\begin{eqnarray}
{\bf V}_1 &=& [0,0,0]^T \\
{\bf V}_2 &=& [1,0,0]^T \\
{\bf V}_3 &=& [1/2, \sqrt {3}/2,0]^T 
\end{eqnarray}

{\bf Location 3: One corner and the median of the triangle}
\begin{eqnarray}
{\bf V}_1 &=& [0,0,0]^T \\
{\bf V}_2 &=& [\sqrt{3}/2, 1/2,0]^T \\
{\bf V}_3 &=& [\sqrt{3}/2,-1/2,0]^T
\end{eqnarray}

With the first location, the singular configuration surface is simple to express but the position depends on the orientation. With the second location, the coordinates of $P$ do not depend on the orientation. However, we lost the symmetrical property of the mobile platform and the singularity surface is more complex. We have selected the third location for further analysis as we are able to solve the DKP and IKP by solving quadratic equations and by a proper change of variables, study the joint space and workspace in a three dimensional space.

The robotics community generally  uses  the Euler or the Tilt-and-Torsion angles to represent the orientation of the mobile platform. These methods have physical meaning, but there exist singularities for representing some orientations.
If $\bf u$, $\bf v$ and $\bf w$ are three unit vectors defined along the axes of moving frame, then the rotation matrix $\bf R$ can be expressed in terms of the direction cosines of $\bf u$, $\bf v$ and $\bf w$ as:
\begin{equation}
{\bf R}= 
  \left[ \begin {array}{ccc} 
    {u_x}&{v_x}&{w_x} \\ 
    {u_y}&{v_y}&{w_y}\\ 
  	{u_z}&{v_z}&{w_z}
  \end {array} \right]
\end{equation}

The unit quaternions give us a redundant way to define the orientation but gives a single definition of any orientations.

\begin{equation}
{\bf R}=
 \left[ \begin {array}{ccc} 
  2 q_1^2+2 q_2^2-1 & -2 q_1 q_4+2 q_2 q_3 & 2 q_1 q_3 + 2 q_2 q_4  \label{EQ:r}\\ 
  2 q_1 q_4+2 q_2 q_3 & 2 q_1^2 + 2 q_3^2-1 & -2 q_1 q_2+2 q_3 q_4 \\ 
 -2 q_1 q_3+2 q_2 q_4 & 2 q_1 q_2+2 q_3 q_4 & 2 q_1^2 + 2 q_4^2-1
\end {array} \right] 
\end{equation}

\noindent with $q_1 geq 0$. To simplify the equations, we can also write the coordinates of the moving platform in the fixed reference frame with general rotation matrix as
\begin{equation}
  {\bf W}_i = {\bf R} {\bf V}_i + {\bf P} \quad {\rm where } \quad {\bf P}= [x, y, z]^T
\end{equation}
with
\begin{eqnarray}
{\bf W}_1 &=& [x,y,z]^T \\
{\bf W}_2 &=& \left[
\begin{array}{c}
u_x \sqrt{3}/2+v_x/2+x \\
u_y \sqrt{3}/2+v_y/2+y \\
u_z \sqrt{3}/2+{v_z}/2+z
\end{array}
\right] \\
{\bf W}_3 &=& \left[
\begin{array}{c}
u_x \sqrt{3}/2-v_x/2+x \\
u_y \sqrt{3}/2-v_y/2+y \\
u_z \sqrt{3}/2-{v_z}/2+z
\end{array}
\right]
\end{eqnarray}
\subsection{Constraint equations}
To solve the DKP, there are two main methods to express the constraint equations. 
For the first method \cite{Parenti-Castelli:1990}, is to find the location of the mobile platform, by looking for the value of the passive prismatic joints $[ \rho_{1x}, \rho_{1y}, \rho_{1z}]$. The distances between points $C_i$ are
\begin{equation}
||{\bf C}_1-{\bf C}_2|| = ||{\bf C}_1-{\bf C}_3|| = ||{\bf C}_2-{\bf C}_3|| = 1
\end{equation}
which can be written as follows :
\begin{equation}
      \begin{aligned}
\left(\rho_{1x}+\rho_{2y} \right)^{2}+ \left(\rho_{1y}-\rho_{2x} \right)^{2}+ \left(\rho_{1z}-\rho_{2z} \right)^{2}&=&1 \label{EQ:C1}\\
\left(\rho_{2y}+\rho_{3y} \right)^{2}+ \left(\rho_{2x}+\rho_{3x} \right)^{2}+ \left(\rho_{2z}-\rho_{3z} \right)^{2}&=&1 \\ 
\left(\rho_{1x}+\rho_{2y} \right)^{2}+ \left(\rho_{1y}-\rho_{2x} \right)^{2}+ \left(\rho_{1z}-\rho_{2z} \right)^{2}&=&1 \\
      \end{aligned}
\end{equation}

This method is also used by \cite{Chen:2012} for the 3-\underline{PP}PS. Finally we have to solve a fourth degrees polynomial equations with complicated coefficients. The constraint equations seem to be simple as shown in Eq.~(\ref{EQ:C1}), but no trivial way exist to have simple analytic solution. When all the lengths of the prismatic joints are know, it is easy to write the orientation of the mobile platform by using any representation.

The second method is to remove the passive joints from the constraint equations. By using the general representation of the orientation, we have the following equations for the passive joints. 
\begin{equation}
      \begin{aligned}
          \rho_{1x} &=& x \\
          \rho_{2x} &=& u_y \sqrt {3}/2+ v_y/2+y \\
          \rho_{3x} &=&-u_y \sqrt {3}/2+ v_y/2-y
      \end{aligned}
\end{equation}

Finally, the constraint equations of the 3-\underline{PP}PS robot are
\begin{equation}
      \begin{aligned}
        \rho_{1y}-y &=& 0 \label{EQ:y}\\
        \rho_{1z}-z &=& 0  \\
       -\rho_{2y}-u_x \sqrt{3}/2-v_x/2-x &=& 0 \\
        \rho_{2z}-u_z \sqrt{3}/2-v_z/2-z &=& 0 \\
        \rho_{3y}-u_x \sqrt{3}/2+v_x/2-x &=& 0 \\
        \rho_{3z}-u_z \sqrt{3}/2+v_z/2-z &=& 0 
      \end{aligned}
\end{equation}

And, by using the rotation matrix (\ref{EQ:r}) with quaternion parameters, substituting in (\ref{EQ:y}), we obtain
\begin{equation}
      \begin{aligned}
        \rho_{1y}-y &=& 0 \label{EQ:yq}  \\
        \rho_{1z}-z &=& 0  \\
        \left( -2 q_1^2-2 q_2^2+1 \right) \sqrt{3}/2 + q_1 q_4 - q_2 q_3-x- \rho_{2y} &=& 0  \\
        \sqrt{3} (q_1 q_3- q_2 q_4) - q_1 q_2 - q_3 q_4 + \rho_{2z}-z &=& 0  \\
        \left( -2 q_1^2-2 q_2^2+1 \right) \sqrt{3}/2 - q_1 q_4 + q_2 q_3-x+ \rho_{3y} &=& 0  \\
        \sqrt{3} (q_1 q_3- q_2 q_4) + q_1 q_2 + q_3 q_4 + \rho_{3z}-z &=& 0  
      \end{aligned}
\end{equation}

The system of equation  (\ref{EQ:yq}) becomes algebraic if we add a parameter $s_3$ with $s_3^2=3$ and substitute $\sqrt{3}=s_3$. The resolution of this system of constraint equations will be done in the Section 4.
\subsection{Change of variables}
We introduce a change of variables to reduce the joint space dimension from six to three where three coordinates are equal to zero. 
\begin{equation}
      \begin{aligned}
  \mu_{1x} &=& \rho_{1x} - \rho_{2y} \label{EQ:c}  \\
  \mu_{1y} &=& \rho_{1y} - \rho_{1y} = 0 \\
  \mu_{1z} &=& \rho_{1z} - \rho_{1z} = 0 \\
  \mu_{2x} &=& \rho_{2x} - \rho_{1y}  \\
  \mu_{2y} &=& \rho_{2y} - \rho_{2y} = 0 \\
  \mu_{2z} &=& \rho_{2z} - \rho_{1z}  \\
  \mu_{3x} &=& \rho_{3x} + \rho_{2y}  \\
  \mu_{3y} &=& \rho_{3y} - \rho_{2y}  \\
  \mu_{3z} &=& \rho_{3z} - \rho_{1z}   
      \end{aligned}
\end{equation}

With this change of variables, the coordinates of the mobile platform are also translated but its orientation does not change.
\begin{eqnarray}
x'&=& x - \rho_{2y} \nonumber\\
y'&=& y - \rho_{1y} \\
z'&=& z - \rho_{1z} \nonumber
\end{eqnarray}
This simplification is similar to the case where the three first prismatic joints are orthogonal as in \cite{Caro:2012}.
\section{Singularity Analysis}
The singular configurations of a parallel robot can be found by writing the serial and parallel Jacobian matrices \cite{Gosselin:1990,Sefrioui:92,Chablat:1998}. These two matrices are obtained by differentiating with respect to time the constraint equations. These serial and parallel Jacobian matrices satisfy the following  relationship
\begin{equation}
  {\bf A t + B \dot{\mathbf{\rho}}}=0
\end{equation}

\noindent where $\bf t$ is the twist of the moving platform and ${\bf \dot{\mathbf{\rho}}}$ is the vector of the active joint velocities.

The topology of the legs of the 3-PPPS robot means that there is no serial singularity because the determinant of the matrix ${\bf B}$ does not vanish. In using the same approach that in \cite{Caro:2010}, we can evaluate the matrix ${\bf A}$ and its the determinant can be factorized as
\begin{equation}
 (q_1^2-q_2^2-q_3^2+q_4^2) (q_1^2-q_2^2+q_3^2-q_4^2)=0
\end{equation}

\noindent We can also remove $q_1$ 
\begin{equation}
  \left( q_2^2+ q_3^2-1/2 \right) \left( q_2^2+ q_4^2-1/2 \right) =0
\end{equation}
with the properties of the quaternion $q_1^2+q_2^2+q_3^2+q_4^2 \leq 1$.

\begin{figure}[!ht]
\begin{center}
\includegraphics[scale=.42]{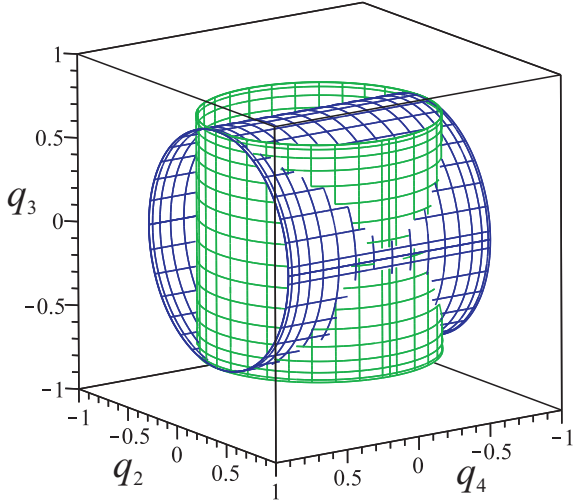}
\end{center}
\caption{Parallel singularity of the 3-\underline{PP}PS robot with quaternion representation}
\label{Fig:Singularity}
\end{figure}

Both equations represent a cylinder whose diameter is $1/2$. Figure~\ref{Fig:Singularity} depicts these surfaces bounded by the unit sphere.
\section{Direct Kinematics}
Generally, finding the solutions of the IKP is simple for parallel robots, whereas,  finding the solutions of the DKP is a complex problem. For a Gough-Stewart platform, we can find up to 40 real solutions \cite{Husty:1996,Innocenti:2001,Merlet:2006}. For the 3-\underline{P}PS, several methods exist to solve the DKP \cite{Parenti-Castelli:1990}. Generally, we obtain a fourth degree polynomial with huge coefficients. 
For our design, it is trivial to have $y$ and $z$ from Eq.~(\ref{EQ:y}).
\begin{eqnarray}
  y&=& \mu_{1y} \nonumber \\
	z&=& \mu_{1z} \nonumber
\end{eqnarray}
If we apply the change of variable (\ref{EQ:c}) in the constraint equation (\ref{EQ:yq}), we obtain.
\begin{equation}
      \begin{aligned}
-\sqrt {3} (q_1^2+ q_2^2)+\sqrt {3}/2+ q_1 q_4-q_2 q_3&=&x  \\
\left(\sqrt {3} q_1-q_4 \right) q_3-\sqrt{3} q_2 q_4- q_1 q_2+ \mu_{2z}&=&0  \\
\mu_{3y}-\sqrt {3} (q_1^2+ q_2^2)+\sqrt {3}/2- q_1 q_4+ q_2 q_3&=&x  \\
\left( \sqrt{3} q_1+q_4 \right) q_3-\sqrt{3}q_2 q_4 + q_1 q_2+\mu_{3z}&=&0  \\
q_1^2+q_2^2+q_3^2+q_4^2&=&1 
\label{eq:Constraint_plus}
      \end{aligned}
\end{equation}
Using Gr\"{o}bner bases, we can eliminate orientation variables to obtain an equation depending on the articular coordinates and position $x$.
So, $x$ is the solution of the following  quadratic equation:
\begin{eqnarray}
&&\left( 4 \mu_{2z}^2-8 \mu_{3z} \mu_{2z} + 4 \mu_{3z}^2 -4 \right) {x}^{2} +  \nonumber \\
&&\left( -8 \mu_{3y} \mu_{2z}^{2}+8 \mu_{2z} \mu_{3y} \mu_{3z}+4\mu_{3y} \right) x+   \nonumber \\
&& 4 (\mu_{3y}^{2} \mu_{2z}^{2} - \mu_{2z}^{2}+ \mu_{3z} \mu_{2z}- \mu_{3y}^{2}-  \mu_{3z}^{2})+3 = 0
\end{eqnarray}
The two roots of this equation are
\begin{eqnarray}
\!&x&=
{\frac{\mu_{3y} \left( 2\mu_{2z}^2-2\mu_{3z}\mu_{2z}-1 \right)}{2 \left(\mu_{2z}-\mu_{3z}+1\right) 
\left(\mu_{2z}-\mu_{3z}-1\right)}} \\
\! &\pm& \! \frac{\sqrt{\left( 4(\mu_{2z}^2- \mu_{3z} \mu_{2z}+ \mu_{3z}^{2})-3 \right)  
\left( (\mu_{2z}-\mu_{3z})^2 +\mu_{3y}^{2}-1 \right) }}
{ 2 \left( \mu_{2z}-\mu_{3z}+1 \right)  \left( \mu_{2z}-\mu_{3z}-1 \right) } \nonumber
\end{eqnarray}
Where singular locus are easily found when $|\mu_{2z}-\mu_{3z}|=1$, $(\mu_{2z}-\mu_{3z})^2 +\mu_{3y}^{2}=1$ and $4(\mu_{2z}^2- \mu_{3z} \mu_{2z}+ \mu_{3z}^{2})=3$.

A similar method is used to find $q_1$. We have the following biquadratic equation
\begin{eqnarray}
&&48 q_1^4 + (16 \sqrt{3} x-24 -8 \sqrt{3} \mu_{3y}) q_1^2 \nonumber \\
&&\left( 4 \sqrt{3}\mu_{3y} - 4 \sqrt {3} x+7 \right) \mu_{2z}^{2} + \left( 7 -4 \sqrt{3} x\right) \mu_{3z}^{2} -3 + \\ 
&&\left( \left( 8 \sqrt{3} x-10 \right) \mu_{3z} -4 \sqrt{3} \mu_{3y} \mu_{3z}\right) \mu_{2z} +  4 (\mu_{3y}^{2} - \mu_{3y}  x  + x^{2})  = 0  \nonumber
\end{eqnarray}
To simplify the writing of the roots, we write the discriminant $\Delta$
\begin{eqnarray}
\Delta_1&=& -9 \mu_{3y}^2-12 \sqrt{3} \mu_{2z}^2 \mu_{3y} + 12 \sqrt{3} \mu_{2z}^{2} x+ \nonumber \\
&&12 \sqrt{3} \mu_{2z} \mu_{3y} \mu_{3z} -24 \sqrt{3} \mu_{2z} \mu_{3z} x + 12 \sqrt{3} \mu_{3z}^2 x +  \\
&&6 \sqrt{3} \mu_{3y} - 12 \sqrt{3} x -21 \mu_{2z}^2 + 30 \mu_{3z} \mu_{2z} - 21 \mu_{3z}^{2} + 18 \nonumber
\end{eqnarray}
Finally, the four roots of $q_1$ are simply written as a function of each $x$
\begin{equation}
q_1= \pm \frac{\sqrt {48 \sqrt{3}\mu_{3y} - 96 \sqrt{3} x + 144 \pm 6 \sqrt {\Delta_1}}
}{24}
\end{equation}
The maximum number of orientations is thus eight, but we must take into account the redundancy of information of the quaternions which doubles the number of possible orientations. To find $q_2$, we obtain a polynomial equation whose coefficients are the same as to find $q_1$, which gives the same discriminant
\begin{equation}
q_2= \pm \frac{\sqrt {48 \sqrt{3}\mu_{3y} - 96 \sqrt{3} x + 144 \pm 6 \sqrt {\Delta_1}}
}{24}
\end{equation}
Similarly, we find a biquadratic equation to find $q_3$ with identical coefficients to that of $q_4$.
\begin{eqnarray}
&& 432 q_3^4 + (72 \sqrt {3} (\mu_{3y} - 2 x) -216) q_3^2 + \nonumber \\
&& \left( -36 \sqrt{3} \mu_{3y} + 36 \sqrt{3} x + 63 \right) \mu_{r2z}^2 + \nonumber \\
&& \left( 36 \sqrt{3} \mu_{3y} \mu_{3z} + \left( -72 \sqrt{3} x- 90 \right) \mu_{3z} \right) \mu_{2z} +  \nonumber \\
&& \left( 36 \sqrt{3} x + 63  \right) \mu_{3z}^2 + 36 \mu_{3y}^2 - 36 \mu_{3y} x + 36 x^2-27 = 0\nonumber 
\end{eqnarray}
and its discriminant
\begin{eqnarray}
&&\Delta_3= 5184 \sqrt{3} \nonumber \\
&&((12 \mu_{3y}-12 x -7 \sqrt{3}) \mu_{2z}^2+(10 \sqrt{3}-12 \mu_{3y}+24 x) \mu_{3z} \mu_{2z}- \nonumber \\
&&(7 \sqrt{3}+12 x) \mu_{3z}^2-3 \mu_{3y}^2 \sqrt{3}+6 \sqrt{3}-6 \mu_{3y}+12 x ) \nonumber
\end{eqnarray}
Finally, we can find out $q_3$ and $q_4$
\begin{eqnarray}
q_3&=&\pm  \frac {\sqrt{-432 \sqrt{3} \mu_{3y} + 864 \sqrt{3} x + 1296 \pm \sqrt{\Delta_3}}}{72} \\
q_4&=&\pm  \frac {\sqrt{-432 \sqrt{3} \mu_{3y} + 864 \sqrt{3} x + 1296 \pm \sqrt{\Delta_3}}}{72}
\end{eqnarray}
With the method described above, we obtain one solution to the DKP for $y$ and $z$, two solutions for $x$ and eight solutions for $q_1$, $q_2$, $q_3$ and $q_4$. Normally all permutations of solutions could be root but we need to verify the following two coupling equations which reduces the number of solutions to the DKP to eight.
\begin{eqnarray}
\sqrt {3} q_1^2+\sqrt{3} q_2^2- q_1 q_4+ q_2 q_3+x &=&\sqrt{3}/2 \nonumber \\
q_1^2+q_2^2+q_3^2+q_4^2&=&1
\label{eq:Couplage}
\end{eqnarray}
Thus, we assume $q_1>0$.
\section{Inverse Kinematics}
Due to the location of the origin on the mobile platform, the computation of the inverse kinematics is simple. The robot admits only one inverse kinematic solution. The result is valid for the general case without the change of variables. Following are the equations associated with the IKP. 
\begin{eqnarray}
\rho_{1y}&=& y  \nonumber \\
\rho_{1z}&=& z  \nonumber \\
\rho_{2y}&=& q_1 q_4 - q_2 q_3 + \sqrt{3}/2 - \sqrt{3}(q_1^2+q_2^2) -x  \nonumber \\
\rho_{2z}&=&  \left( \sqrt{3}q_2 + q_3  \right) q_4 - \left(\sqrt{3} q_3 - q_2 \right) q_1 + z  \nonumber \\
\rho_{3y}&=& q_1 q_4- q_2 q_3 - \sqrt{3}/2 + \sqrt{3}(q_1^2+q_2^2) + x  \nonumber \\
\rho_{3z}&=&  \left( \sqrt{3} q_2 - q_3 \right) q_4 - \left(\sqrt{3} q_3 + q_2 \right) q_1 + z \nonumber 
\end{eqnarray}

\noindent Once this computation is done, we can compute the change of variables to study the joint space.
\section{Workspace and Joint Space Analysis}
By using a CAD \cite{Cbook:75}, it becomes possible to model the workspace and joint space of the 3-\underline{PP}PS robot. This modeling uses a set of cells similar to intervals for which we know the boundaries thanks to the discriminant variety of the constraints equations \cite{Lazart:2007} and the coordinates of a point inside the cell. With this point, it is possible to evaluate, for example, the number of solutions to the DKP. We know that inside a cell the properties of the robot do not change. In this work, we use the SIROPA library which allows spatial representations of cells that do not exist in Maple. To define the aspects, that is to say the maximal domains of the workspace without singularities, we add to the equations of constraints of the robot the components resulting from the factorization of the determinant of $\bf A$ \cite{Chablat:1998}.
\subsection{Joint space analysis}
The study of the joint space with the CAD allows to say that the DKP always admits 16 real roots which corresponds to eight assembly modes for the robot. 
This result is valid if there is no limit on the passive joints.
The CAD gives us two important results which are the discriminant variety and the projection of these polynomials on the axes of representation.

\begin{figure*}[!ht]
\begin{center}
\includegraphics[scale=.7]{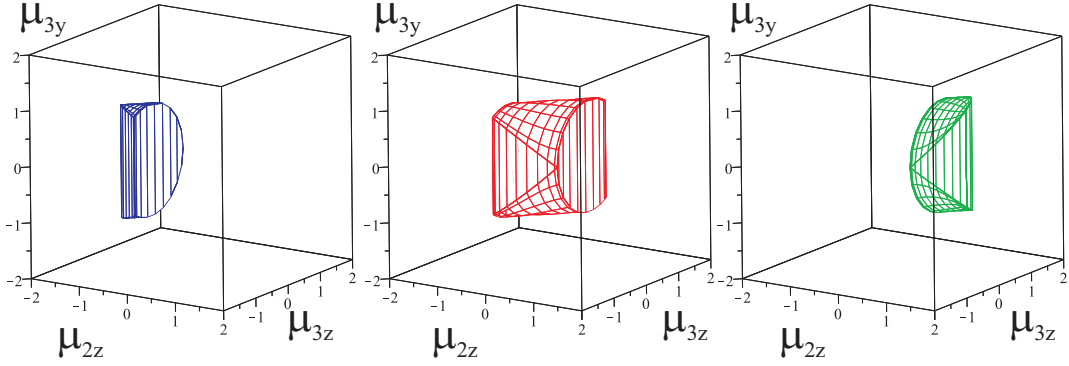}
\end{center}
\caption{Three cells to define the joint space of the 3-\underline{PP}PS robot}
\label{Fig:JointSpace_Cells}
\end{figure*}

\noindent For the joint space, the discriminant variety is
\begin{eqnarray}
\mu_{2z} - \mu_{3z} &=&1 \nonumber \\
\mu_{2z} - \mu_{3z} &=& -1 \nonumber \\
4 (\mu_{2z}^2 -  \mu_{2z} \mu_{3z} +  \mu_{3z}^2) -3 &=& 0 \nonumber \\
(\mu_{2z}- \mu_{3z})^2+\mu_{3y}^2-1 &=& 0 \nonumber 
\end{eqnarray}

\begin{figure}[!ht]
\begin{center}
\includegraphics[scale=.65]{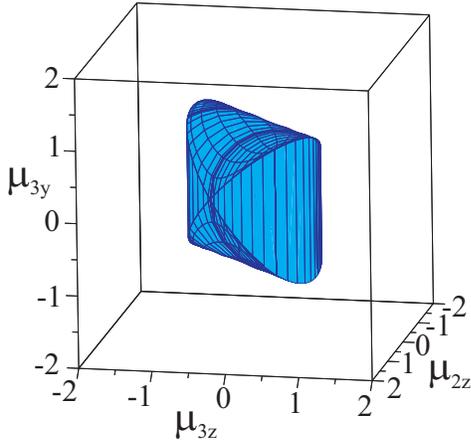}
\end{center}
\caption{Joint space of the 3-\underline{PP}PS robot}
\label{Fig:JointSpace}
\end{figure}

\noindent And the projection ${\cal P}$ of these polynomials are for the $\mu_{2z}$, $\mu_{3z}$ and $\mu_{2y}$ axis
\begin{eqnarray}
{\cal P}_{1_{R_{2Z}}}: \mu_{2z}+1&=&0 \nonumber \\
{\cal P}_{2_{R_{2Z}}}: 2 \mu_{2z}+1&=&0 \nonumber \\
{\cal P}_{3_{R_{2Z}}}: 2 \mu_{2z}-1&=&0 \nonumber \\
{\cal P}_{4_{R_{2Z}}}: \mu_{2z}-1&=&0 \nonumber  \\
%
{\cal P}_{1_{R_{3Z}}}: \mu_{2z}-\mu_{3z}-1&=&0 \nonumber \\
{\cal P}_{2_{R_{3Z}}}: \mu_{2z}-\mu_{3z}+1&=&0 \nonumber \\
{\cal P}_{3_{R_{3Z}}}: 4 (\mu_{2z}^{2}- \mu_{2z} \mu_{3z}+ \mu_{3z}^{2})-3&=&0 \nonumber \\
%
{\cal P}_{1_{R_{3Y}}}: (\mu_{2z}-\mu_{3z})^2+ \mu_{3y}^{2}-1&=&0 \nonumber 
\end{eqnarray}

\noindent In \cite{Moroz:2010}, the cell description is done to explain the Table~\ref{Table:Jointspace_CAD}. 
For one variable, $[{\cal P},n, \mu, {\cal Q},m]$ means that the minimum value of $\mu$ is the $n^{th}$ root of ${\cal P}$ and 
the maximum value is $m^{th}$ root of ${\cal Q}$. Figure~\ref{Fig:JointSpace_Cells} depicts the three cells separately of the joint space and the Figure~\ref{Fig:JointSpace} their assemblies.

\begin{table*}[htb]
    \begin{center}
		{\small
\begin{tabular}{|c|c|c|}
   \hline
   $\mu_{2z} $ &
   $\mu_{3z} $ &
   $\mu_{3y} $ \\ \hline   
   $[{\cal P}_{1_{R_{2Z}}}, 1, \mu_{2z}, {\cal P}_{2_{R_{2Z}}}, 1]$ & 
   $[{\cal P}_{3_{R_{3Z}}}, 1, \mu_{3z}, {\cal P}_{3_{R_{3Z}}}, 2]$ & 
   $[{\cal P}_{1_{R_{3Y}}}, 1, \mu_{3y}, {\cal P}_{1_{R_{3Y}}}, 2]$ \\
   $[{\cal P}_{2_{R_{2Z}}}, 1, \mu_{2z}, {\cal P}_{3_{R_{2Z}}}, 1]$ &
   $[{\cal P}_{3_{R_{3Z}}}, 1, \mu_{3z}, {\cal P}_{3_{R_{3Z}}}, 2]$ & 
   $[{\cal P}_{1_{R_{3Y}}}, 1, \mu_{3y}, {\cal P}_{1_{R_{3Y}}}, 2]$ \\
   $[{\cal P}_{3_{R_{2Z}}}, 1, \mu_{2z}, {\cal P}_{4_{R_{2Z}}}, 1]$ &
   $[{\cal P}_{3_{R_{3Z}}}, 1, \mu_{3z}, {\cal P}_{3_{R_{3Z}}}, 2]$ & 
   $[{\cal P}_{1_{R_{3Y}}}, 1, \mu_{3y}, {\cal P}_{1_{R_{3Y}}}, 2]$ \\ \hline
\end{tabular}}
\end{center}
\caption{Joint space description by three cells from CAD}
\label{Table:Jointspace_CAD}
\end{table*}
\subsection{Workspace analysis}
The singularity analysis allows us to know the locus where the robot reaches parallel singularities. The aim of the analysis is to determine the maximum regions without any singularities, i.e. the aspects of the robot. In these regions, the robot can perform any continuous trajectories. As the determinant of the parallel Jacobian matrix $\bf A$ can be factorized in two components, the orientation space is divided in four regions by using the sign of two components.
\begin{itemize}
 \item Let PP the regions where ${q_2}^{2}+{q_3}^{2}-1/2>0$ and ${q_2}^{2}+{q_4}^{2}-1/2>0$.
 \item Let NN the regions where ${q_2}^{2}+{q_3}^{2}-1/2<0$ and ${q_2}^{2}+{q_4}^{2}-1/2<0$.
 \item Let PN the regions where ${q_2}^{2}+{q_3}^{2}-1/2>0$ and ${q_2}^{2}+{q_4}^{2}-1/2<0$.
 \item Let NP the regions where ${q_2}^{2}+{q_3}^{2}-1/2<0$ and ${q_2}^{2}+{q_4}^{2}-1/2>0$.
\end{itemize}

Each region can be defined by a set of cells. According to the projection axis, the number of cell changes. In the Table~\ref{Table:Cells}, we have the number of cells to define each aspect as a function of the sign of each component of the determinant of $\bf A$. If the same projection axes are used, all aspects are defined by 24 cells. If we use a different order for each aspect, we have only 20 cells.

\begin{figure}[!hb]
\begin{center}
\includegraphics[scale=.5]{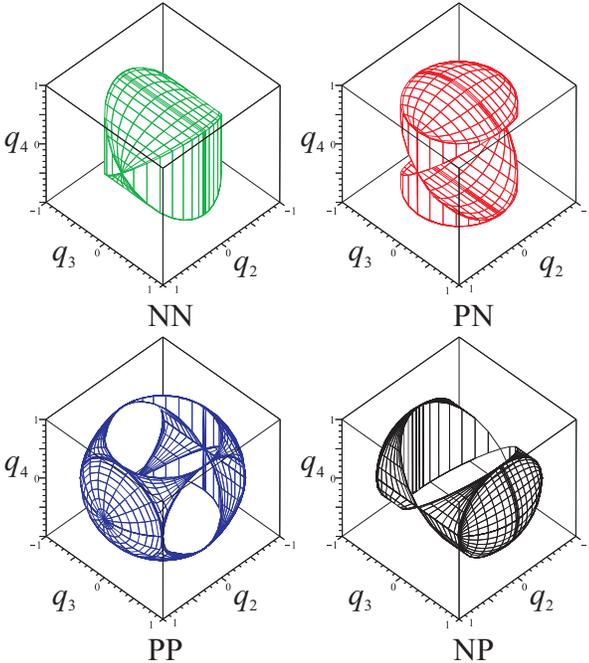}
\end{center}
\caption{Workspace of the 3-\underline{PP}PS robot}
\label{Fig:WorkSpace}
\end{figure}

The discriminant variety of the constraint equations with the singularity condition is
\begin{eqnarray}
 2 q_2^{2}+2 q_3^2-1&=&0 \nonumber\\
 2 q_2^{2}+2 q_4^2-1&=&0 \\
 q_2^2+ q_3^2+q_4^2-1&=&0 \nonumber
\label{eq:Dis}
\end{eqnarray}

The projection ${\cal P}$ into the three axis $q_2$, $q_3$ and $q_4$ are
\begin{eqnarray}
{\cal P}_{q_2}&:& q_2=0,-1+ q_2=0,  q_2+1=0, 2 q_2^2-1 = 0 \nonumber\\
{\cal P}_{q_3}&:& 2 q_3^2-1=0, 2 q_2^2+2 q_3^2-1=0, q_2^2+q_3^2-1 = 0 \\
{\cal P}_{q_4}&:& 2 q_2^2+2 q_4^2-1=0, q_2^2+ q_3^2+ q_4^2-1 = 0 \nonumber
\label{eq:Proj}
\end{eqnarray}

Table~\ref{Table:Workspace_CAD} presents the cell description of the NN aspect with only two cells. This aspect contains the \textit{home} pose of the robot as in Fig.~\ref{Fig:Robot}. For any trajectory described by quaternion, simple test can give us if one posture is inside the NN aspect and the intersection of a parametrization of the trajectory with the boundary equations of this region as in\cite{Jha:2015}. When the boundaries are know, the algebraic tools permit us to project this result in any parametrization able to describes the orientation of the mobile platform.

\begin{table}[!ht]
  \begin{center}
  \begin{tabular}{|c|c|c|c|c|c|}
    \hline
		Order & PP & PN & NN & NP & Total \\ \hline
			$q_2$, $q_3$, $q_4$ & 12 & 12 & 6 & 6 & 36	\\
			$q_2$, $q_4$, $q_3$ & 10 & 8  & 2 & 4 & 24 \\
			$q_3$, $q_4$, $q_2$ & 12 & 12 & 6 & 6 & 36 \\
			$q_3$, $q_2$, $q_4$ & 12 & 6  & 2 & 4 & 24 \\
			$q_4$, $q_2$, $q_3$ & 12 & 4  & 2 & 6 & 24 \\
			$q_4$, $q_3$, $q_2$ & 12 & 12 & 6 & 6 & 36 \\ \hline
  \end{tabular}
  \end{center}
	\caption{Number of cells to model the workspace according to the projection axis order}
	\label{Table:Cells}
\end{table}

\begin{table*}[htb]
    \begin{center}
		{\small
\begin{tabular}{|c|c|c|}
   \hline
   $q_2 $ &
   $q_3 $ &
   $q_4 $ \\ \hline   
   $[2 q_2^2-1,1,q_2, q_2,1] $ &
   $[2 q_2^2+2 q_3^2-1,1, q_3,2 q_2^2+2 q_3^2-1,2]$ &
   $[2 q_2^2+2 q_4^2-1,1, q_4,2 q_2^2+2 q_4^2-1,2]$ \\
   $[q_2,1, q_2,2 q_2^2-1,2]$ &
   $[2 q_2^2+2 q_3^2-1,1, q_3,2 q_2^2+2 q_3^2-1,2]$ &
   $[2 q_2^2+2 q_4^2-1,1, q_4,2 q_2^2+2 q_4^2-1,2]$ \\ \hline
\end{tabular}}
\end{center}
\caption{Modeling of the NN aspect by two cells from the CAD}
\label{Table:Workspace_CAD}
\end{table*}
\section{Conclusions and Perspectives}
In this paper, we have studied the workspace, the singularities, the workspace and the joint space of a 3-\underline{PP}PS parallel robot.  The proposed design  with U-shape base permits to have simpler kinematics, which can be solved in real time. To the knowledge of the authors, it's the first 6-DOF parallel robot where the DKP is solvable with quadratic equations. By the way, in the control loop, we can check with a high frequency, the position of the end-effector according to the joint position as in\cite{Caro:2015}. The appropriate selection of parameters to represent the position and orientation of the mobile platform simplifies the constraint equations.  There are no serial singularities for the proposed 3-\underline{PP}PS parallel robot, however there exists parallel singularities which only depend on the orientation of the end-effector. The workspace and joint space can be easily characterized with a low number of cells by the CAD. The aspects associated with each assembly mode can be represented in a 3D space by changing the coordinates. 
Further works will be to make the embodiment of this architecture and to add the constraints of the joint limits in the workspace and joint space definitions. Finally, the stiffness analysis has to be done in order to evaluate which applications are suitable for this robot architecture. 

\bibliographystyle{asmems4}

\end{document}